# Bayesian crack detection in ultra high resolution multimodal images of paintings

Bruno Cornelis*†, Yun Yang‡, Joshua T. Vogelstein‡§, Ann Dooms*†, Ingrid Daubechies§ and David Dunson‡

*ETRO (Department of Electronics and Informatics), Vrije Universiteit Brussel, Belgium
†iMinds VZW, Gaston Crommenlaan 8 box 102, 9050 Ghent, Belgium
‡Department of Statistical Science, Duke University, Durham, North Carolina, USA
§Department of Mathematics, Duke University, Durham, North Carolina, USA
Email: bcorneli@etro.vub.ac.be

## Abstract

[1] The preservation of our cultural heritage is of paramount importance. Thanks to recent developments in digital acquisition techniques, powerful image analysis algorithms are developed which can be useful non-invasive tools to assist in the restoration and preservation of art. In this paper we propose a semi-supervised crack detection method that can be used for high-dimensional acquisitions of paintings coming from different modalities. Our dataset consists of a recently acquired collection of images of the Ghent Altarpiece (1432), one of Northern Europe's most important art masterpieces. Our goal is to build a classifier that is able to discern crack pixels from the background consisting of non-crack pixels, making optimal use of the information that is provided by each modality. To accomplish this we employ a recently developed non-parametric Bayesian classifier, that uses tensor factorizations to characterize any conditional probability. A prior is placed on the parameters of the factorization such that every possible interaction between predictors is allowed while still identifying a sparse subset among these predictors. The proposed Bayesian classifier, which we will refer to as *conditional Bayesian tensor factorization* or *CBTF*, is assessed by visually comparing classification results with the Random Forest (RF) algorithm.

## Index Terms

Ghent Altarpiece, classification, crack detection, nonparametric Bayes, tensor factorization, variable selection

## I. Introduction

The digitization of cultural artifacts is becoming an increasingly popular practice. Although the main goal is preservation, documentation and dissemination, digital analysis is a key emerging *non-invasive* tool for art historians and preservation experts. In this work, we focus on crack detection in the *Ghent Altarpiece*, dated by inscription 1432. Jan Van Eyck, its painter, is considered to be one of the most talented artists of his generation, although it is common belief that the polyptych was started by his older brother Hubert Van Eyck. The polyptych panel painting is one of the brothers' most important masterpieces and is still located in its original home, the Saint Bavo Cathedral in Ghent, where it continues to attract many thousands of visitors. After an extensive image acquisition campaign, which ran from April 2010 through June 2011, to assess the structural condition of the Ghent Altarpiece, it was decided it needed to be restored at a cost of more than 1 million euros.

---

[1]This is a preprint version, prepared for posting on ArXiv. It incorporates corrections made by the authors in response to comments by reviewers for the DSP 2013 conference, for which this paper has been accepted. It does not incorporate any subsequent editing changes by IEEE (such as page numbers, for instance) made in preparation for the final, published version. It is the authors' understanding that, under the rules posted on http://www.ieee.org/documents/author_faq.pdf, this posting does therefore not infringe on subsequent copyright transfer to IEEE. When the DOI of the published version will be communicated to us, we will post a link to this as the next version of this paper.

Being able to accurately detect cracks is of major importance in the context of art conservation since cracking is one of the most common forms of deterioration found in paintings. Cracks are a sign of the aging of the materials and are a record of the painting's degradation. Generally speaking, a crack appears in paint layers when pressures develop within or on it because of the influence of various factors and cause the material to break [1]. Typically, for most $15^{th}$ century Flemish paintings on Baltic oak, humidity fluctuations cause the wooden support to shrink or expand and are the main cause for crack formation. These age related or mechanical cracks can affect the entire paint layer structure, including the preparation layer. Premature cracking on the other hand originates in only one of the layers of paint and generally reveals a flawed technical execution during painting, such as not leaving enough time for a layer to dry. A third type of cracks form only in the varnish layer when it becomes brittle through oxidation and are called varnish cracks [1].

Digital image processing can be used to automatically detect crack-like patterns. In the literature, the process of detecting such elongated structures is usually referred to as *ridge-valley structure extraction* [2]. When observing photographs of paintings, cracks can roughly be categorized in two classes, bright cracks on a dark background or dark cracks on a bright background. Generally dark cracks are treated in the literature, where they are typically considered as having low luminance and being local (grayscale) intensity minima with elongated structure [3]. An overview of different crack detection techniques can be found in [4]. These include different types of thresholding, the use of multi-oriented filters (such as Gabor filters) and a plethora of morphological transforms. These methods are also used in other domains for the detection of vessels in medical imagery, roads in satellite imagery and structural damage in manufacturing and engineering (e.g. the detection of cracks in pavement [5]).

The cracks considered here are challenging in a number of ways and require a new set of methods. The crack width ranges from very narrow and barely visible to more significant lacunas. Furthermore, some of the brushstrokes are of similar color and structure as the cracks, which makes their detection daunting. Previous efforts on crack detection in the Ghent Altarpiece can be found in [6] and an extension in [7], where three detection methods (filtering with elongated filters, a multi scale top hat transform and K-SVD for crack enhancement) are proposed. The thresholded outputs of these methods are subsequently combined by using a voting scheme. The Van Eyck images that we analyzed back then are high resolution scans of original photographic negatives, with the capturing process undocumented and the images quite noisy. However, thanks to the project *Lasting Support* (2010-2011), a new *multimodal* dataset was made available in February 2012. The painted surfaces of the altarpiece were documented with digital macrophotography in both the visible and infrared (IR) parts of the electromagnetic spectrum, with previously acquired X-rays also made available.

Each modality has its specific advantage. The infrared photographs can reveal underdrawings and X-rays can reveal changes between earlier paint layers and the final surface. Their high penetration also provides valuable information about structural aspects of the paintings, such as the wood grain and splits in the oak support panels, and about cracks

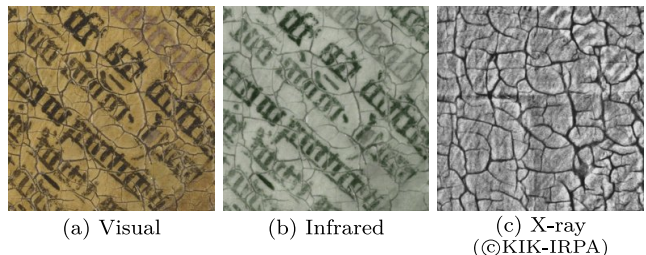

(a) Visual   (b) Infrared   (c) X-ray
(©KIK-IRPA)

Fig. 1: Example of all available modalities

and losses, found in all paint layers. An example of each modality is depicted in Fig. 1, which shows one small portion of the painting containing words in a book.

Since the new dataset consists of a collection of different modalities, including ultra high-resolution images, new crack detection techniques are needed that are able to combine the information provided by each modality. Simply applying the methods from [7], which were designed to work well on one single modality, would require choosing an additional set of parameters per modality. In the new methodology a first step consists in spatially aligning (i.e. registering) all modalities. Once this is achieved, a large feature (i.e. predictor) vector is extracted for each pixel in the image, which results in a high dimensional collection of categorical predictor vectors. The goal now is to build an accurate classifier and select a subset of most important predictors. We define a model, by using a carefully-structured *Tucker factorization*, that can characterize any conditional probability and facilitates variable selection and the modeling of higher-order interactions [8]. We follow a Bayesian approach and use a Markov chain Monte Carlo (MCMC) algorithm for posterior computation and hence allow for uncertainties in the to be included predictors. We compare our results with the random forest (RF) algorithm [9], which is considered as being the best of many competitors and state of the art. The proposed nonparametric Bayesian approach (which we refer to as *conditional Bayesian tensor factorization* or *CBTF*) shows comparable detection results to RF but has the tendency to produce finer crack maps and is faster, which is an attractive property when working with ultra high-dimensional data.

## II. Data preprocessing

In order to be able to use all modalities for crack detection, an initial registration step is required. The images were already approximately registered in order to view them adjacently on the *Closer To van Eyck* website[2]. Unfortunately the precision of this alignment is insufficient in the context of crack detection since cracks are structures that are at most a couple of pixels wide. The modalities of the painting are fundamentally very different and therefore direct registration is a challenging task. However, a more or less consistent component in all modalities are the cracks themselves. Therefore, crude crack maps are initially constructed. These are obtained by first filtering the image with elongated filters such that cracks are enhanced in the image. A subsequent thresholding step produces binary *crack maps* that show an initial estimate of the crack locations. More details on how these elongated filters are constructed and on the particular type of thresholding used are described in [7]. Once the crack maps are obtained for the X-ray and the IR images (used as the reference image), they are registered using the algorithm described in [10]. The transformation that is determined for the registration of the crack maps is then subsequently applied to the original images.

[2]Project website: http://closertovaneyck.kikirpa.be/
Digital image processing: iMinds-ETRO-VUB
Website development: Universum Digitalis

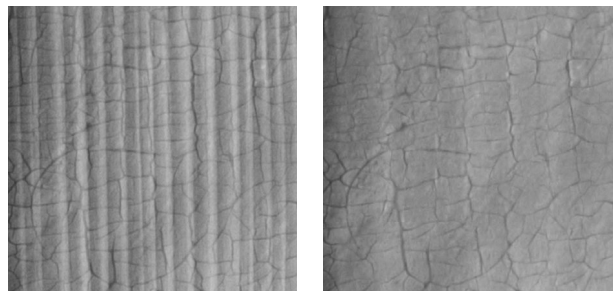

(a) Original      (b) Piecewise smooth component

Fig. 2: Example of MCA decomposition

### A. Enhancing X-rays

The penetrating nature of X-rays make them an appropriate modality for the detection of cracks. Unlike paint that contains lead white, the pigment most opaque to X-rays, cracks appear very dark and are usually clearly delineated. However, the presence of the wood grain in some of the panels can act as a confounding factor during the detection process. In [11], a method called morphological component analysis (MCA) was proposed to separate the texture from the natural part in images. MCA constructs a sparse representation of a signal or an image considering that it is a combination of features which are sparsely represented by different dictionaries. The textural part in our images consists of the wood grain, which is very periodic and suggests the use of a Discrete Cosine Transform (DCT) as one of the dictionaries. The two dictionaries we select are the (Local) Discrete Cosine Transform dictionary, appropriate for a sparse representation of either smooth or periodic behavior, and the Dual Tree Complex Wavelet dictionary, which is appropriate for the representation of piecewise smooth content. An example of such a decomposition is depicted in Fig. 2.

Additionally, some of the panels in the Ghent Altarpiece were reinforced with wooden beams on the back to counteract the moving of the wooden support. These beams appear as a bright grid on the X-rays since the radiation is attenuated more at those places. To cancel this effect we first blur the image, subtract the minimum gray value from all pixels and subtract this blurred version of the image from the original. The resulting image is then contrast enhanced with the method described in [12]. The result of such an operation can be observed in Fig. 3.

### B. Image features

Each modality is processed and filtered with a wide range of filters commonly used in image processing. The methods and filters that are used to construct the image predictor vectors are listed and briefly described below:

- *Elongated filters*: Oriented elongated filters were originally introduced to detect and enhance blood vessels of different thicknesses and orientations in medical images [13]. The filters are made directional by performing weighted linear combinations of partial derivatives of 2D Gaussian kernels (with standard deviations $\sigma = 1$ and 2).
- *Frangi Vesselness filter* [14]: We approximate the Hessian of the image by filtering it with the second derivative of a Gaussian. The Hessian is used with the purpose of developing a vessel enhancement filter. A vesselness measure is obtained for each pixel by analyzing the eigenvalues of the Hessian.

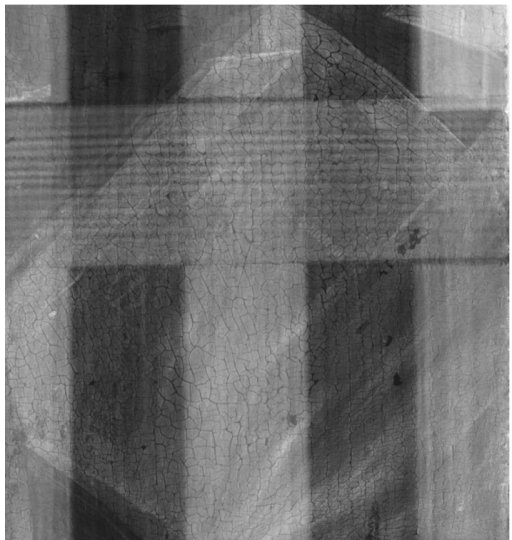
(a) After MCA

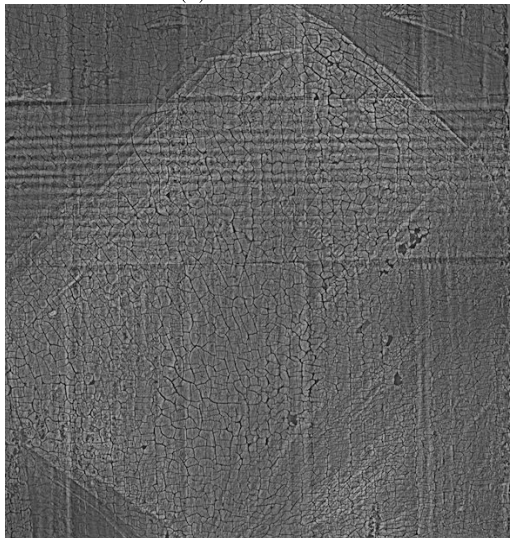
(b) Contrast enhanced

Fig. 3: Enhanced X-ray (©KIK-IRPA, Brussels)

- *Structure tensor* [15]: The structure tensor is a matrix derived from the gradient of an image. It gives information about the predominant directions of the gradient in a specified neighborhood of a point, and the degree to which those directions are coherent. Eigen-decomposition is applied to the structure tensor matrix to form the eigenvalues ($L1, L2$) and eigenvectors ($e1, e2$) respectively. These gradient features provide a more precise description of the local gradient characteristics.
- *Black Top Hat transform*: A popular technique to detect details with particular sizes is the use of a morphological filter known as the *top-hat transformation* [16] which was already successfully applied in crack detection [3], [17], [18]. For the detection of dark cracks on a lighter background generally the *black top-hat* (or *closing top-hat*) transform is used. It is defined as the difference between the morphological *closing* of a grayscale image by a structuring element and the input image and results in a grayscale image with enhanced details. The structuring elements are of size $2 \times 2$ and $3 \times 3$.

- *Local Binary Patterns* [19]: The LBP texture analysis operator allows for detecting particular local binary patterns in circular neighborhoods at a specific spatial resolution. For each pixel in the image, a binary code is produced by thresholding the interpolated pixel values of the circular neighborhood with the value of its center pixel. The most important properties of the LBP operator are its tolerance against illumination changes (i.e. grayscale invariance) and its rotation invariance. The LBP was extended here to react to patterns corresponding to crack-like shapes and assigns a categorical value depending on the detected binary pattern. The size of the neighborhood consists of 16 interpolated grayscale values at a radius of 3 pixels.
- *Color/grayscale intensity*: Cracks have a particular color distribution which can be used to improve the detection rate. Especially in X-rays, cracks appear consistently dark.
- *Median Filter*: The median filter is a nonlinear digital filtering technique, particularly effective in the removal of spiky noise. Three filter sizes are considered: $3 \times 3$, $6 \times 6$ and $12 \times 12$.
- *Laplacian of Gaussian (LoG)*: The LoG filter is typically used to highlight regions of strong and abrupt intensity changes and is therefore often used for edge detection. The standard deviations $\sigma$ that are chosen: 1, 2 and 5.
- *Filter Banks*: The filter bank used in this application is referred to as the *Leung-Malik* filter bank, a multi-scale, multi-orientation filter bank with 48 filters. It consists of first and second derivatives of Gaussians at 6 orientations and 3 scales, 8 LoG filters and 4 Gaussians.

Each pixel is characterized by a feature vector that is the result of all the operations described above, applied to each modality independently. This feature vector sums up to 208 features for each pixel. Once these are extracted they are quantized into an experimentally chosen number of bins $d$ and used in the subsequent classification step, described below.

## III. Bayesian Conditional Tensor Factorizations for High-Dimensional Classification

There is a vast literature on methods for prediction and variable selection from high or even ultra high-dimensional features with a categorical response. Here, we use a recently developed framework for nonparametric Bayes classification through tensor factorizations of the conditional probability $P(Y|X)$, with $Y$ a categorical response and $X = (X_1, \ldots, X_p)$ a vector of $p$ categorical predictors [8]. In our case the values of $Y$ are either 0 or 1, meaning a pixel belongs to the background or it is part of a crack. The proposed method, which is called *conditional Bayesian tensor factorization* or *CBTF*, is based on the principle that a conditional probability can be expressed as a multidimensional array, i.e. a $d_1 \times \cdots \times d_p$ dimensional tensor $A = [a_{i_1, \cdots, i_p}]_{d_1 \times \cdots \times d_p}$, with $d_j$, the number of quantization bins of the $j$th predictor $X_j$ and

$$a_{i_1, \cdots, i_p} = P(Y = 1 | X_1 = i_1, \cdots, X_p = i_p).$$

This *conditional probability tensor* has non-negative cells, i.e. $a_{i_1, \cdots, i_p} \geq 0$ for all $i_1, \ldots, i_p$. To determine the tensor values we perform a low rank tensor factorization where the resulting coefficients correspond to latent class allocation probabilities and response class probabilities.

The literature on tensor factorizations focuses on two types of generalizations of the singular value decomposition (SVD) of matrices. Recall that SVD factorizes a matrix $A = [a_{i_1,i_2}]_{d_1 \times d_2}$ as $A = USV^T$ where $U$ and $V$ are orthogonal matrices and $S$ is a diagonal matrix, thus:
$$a_{i_1,i_2} = \sum_{j_1} \sum_{j_2} s_{j_1,j_2} u_{i_1,j_1} v_{i_2,j_2}.$$

The most popular tensor factorization method is parallel factor analysis (PARAFAC), which expresses a tensor as a sum of $r$ rank-one tensors. The second approach is known as *Tucker decomposition*, which was initially proposed by [20] as a decomposition method for three-way arrays. This was later extended for arrays of arbitrary orders and termed *multilinear SVD* or *higher-order singular value decomposition* (HOSVD) in [21]. Practically, computing an HOSVD of an $n$th-order tensor leads to the computation of $n$ different matrix SVDs. The HOSVD decomposes the tensor $A = [a_{i_1,\cdots,i_p}]_{d_1 \times \cdots \times d_p}$ as:

$$a_{i_1,\cdots,i_p} = \sum_{j_1=1}^{d_1} \cdots \sum_{j_p=1}^{d_p} s_{j_1,\cdots,j_p} \prod_{m=1}^{p} u_{i_m,j_m}^{(m)},$$

where all $U^{(m)} = [u_{i,j}^{(m)}]_{d_m \times d_m}$ are orthogonal matrices and $S = [s_{j_1,\cdots,j_p}]_{d_1 \times \cdots \times d_p}$ is a so called all-orthogonal *core* tensor.

In [8] it was shown that any conditional probability tensor can be decomposed in a similar (non-unique) fashion

$$P(Y = y | X_1 = x_1, \cdots, X_p = x_p)$$
$$= \sum_{j_1=1}^{k_1} \cdots \sum_{j_p=1}^{k_p} \lambda_{j_1,j_2,\cdots,j_p}(y) \prod_{m=1}^{p} \pi_{j_m}^{(m)}(x_m), \qquad (1)$$

where moreover

$$\sum_{j_m=1}^{k_m} \pi_{j_m}^{(m)}(x_m) = 1, \qquad (2)$$

holds for every combination of $(m, x_m)$ and where we assume that the values of $k_m \in \{1, \cdots, d_m\}$ are chosen as small as possible. The factorization coefficients $\lambda_{j_1,j_2,\cdots,j_p}(y)$ can be seen as the latent class allocation probabilities and $\pi_{j_m}^{(m)}(x_m)$ as the response class probabilities and are non-negative.

A primary goal is reducing the enormous amount of covariates, as we expect that a vast majority of the possible combinations between $Y$ and $X$ are negligible (i.e. only a few features will have a significant impact on the classification results), by imposing sparsity. The $k_m$ value impacts the number of parameters used to characterize the $m$th predictor as well as sparsity. In the special case in which $k_m = 1$ and when taking into consideration the constraints in (2), this results in $\pi_1^{(m)}(x_m) = 1$, which means that $P(y|x_1,\cdots,x_p)$ will not depend on $x_m$ and the $m$th predictor is excluded from the model. In other words, sparsity can be imposed by setting $k_m = 1$ for most $m$'s. Additionally, $k_m$ can be seen as the number of latent classes for the $m$th covariate. The levels of $X_m$ are clustered according to their relationship with the response variable in a soft-probabilistic manner, where $k_m$ controls the complexity of the latent structure.

To complete our Bayesian model we choose independent *Dirichlet* priors (commonly used in Bayesian statistics) for the parameters $\Lambda = \{\lambda_{j_1,\cdots,j_p}; j_m = 1, \cdots, k_m; m =$ $1, \cdots, p\}$ and $\pi = \{\pi_{j_m}^{(m)}(x_m); j_m = 1, \cdots, k_m; x_m = 1, \cdots, d_m; m = 1, \cdots, p\}$,

$$\{\lambda_{j_1,\cdots,j_p}(0), \lambda_{j_1,\cdots,j_p}(1)\} \sim Dirichlet(1/2, 1/2),$$
$$\{\pi_1^{(m)}(x_m), \cdots, \pi_{k_m}^{(m)}(x_m)\} \sim Dirichlet(1/k_m, \cdots, 1/k_m).$$

These priors have the advantages of imposing the non-negativity and summing to one constraints. The hyper parameters in the Dirichlet priors are chosen to favor placing most of the probability on a few elements, including near sparsity in these vectors. To embody our prior belief that only a small number of $k_j$'s are greater than one we set

$$P(k_j = 1) = 1 - \frac{r}{p}, \ P(k_j = k) = \frac{r}{(d_j - 1)p},$$

for $k = 2, \cdots, d_j$, $j = 1, \cdots, p$, and where $r$ is the expected number of predictors to be included. To further impose sparsity, we include a restriction that $\#\{j : k_j > 1\} \leq \bar{r}$, where $\bar{r}$ is a prespecified maximum number of predictors. The effective prior on the $k_j$'s is

$$P(k_1 = l_1, \cdots, k_p = l_p)$$
$$= P(k_1 = l_1) \cdots P(k_p = l_p) I_{\{\#\{j:k_j>1\}\leq\bar{r}\}}(l_1, \cdots, l_p),$$

where $I_A(\cdot)$ is the indicator function for set $A$. The full conditional posterior distributions of $\Lambda$ and $\pi$ are of the same family as the prior distributions.

The representation (1) is many-to-one and the different parameters in the factorization cannot be uniquely identified. This does not hamper our Bayesian approach and indeed over-parameterized models often have computational advantages in leading to simplified posterior computation and reduced autocorrelation in Markov Chain Monte Carlo (MCMC) samples of parameters of interest.

The BCTF classifier has a number of appealing properties. There is no parameter tweaking involved which is usually a time consuming process when working with more common image processing operations. The second interesting fact is that the output of the BCTF is actually the probability for each pixel of being a crack pixel. This means that we can propose a *probability crack map* instead of a *binary crack map* as a solution (both are depicted in Fig. 4). A binary crack map is obtained by thresholding the corresponding probability map. A pixel is considered to be part of a crack when it has a crack probability of 0.5 or higher. The fact that the proposed BCTF approach makes full use of the multimodal nature of our dataset results in fewer false positives when compared to our previously developed methods in [7], where only noisy scans were available. Moreover, the BCTF method is significantly faster than the RF classifier, a patch of size $256 \times 256$ is processed within 10 seconds with BCTF while it takes 18 seconds on average with a RF classifier (measured in Matlab 2012a on a laptop with a 2.66Ghz Intel Core i7 processor).

## IV. Experiments

Our dataset consists of a selection of historically important objects in the polyptych. One example is a book in the *Annunciation to Mary* panel. The text depicted in the book, written in a so-called *littera formata*, has puzzled art historians for years. Some words in the text were already deciphered, like the words *De visione Dei* (on the vision of God). Since the Altarpiece is not accessible directly (it is being kept inside a vault behind glass) the text can only be studied from high resolution photographs. Additionally, the presence of cracks in these images makes the book quite difficult to decipher. Thanks to the digital restoration methods of [7] new words were discovered successfully. Obviously, the quality of the digital restoration of the image, done by inpainting, strongly depends on the correct

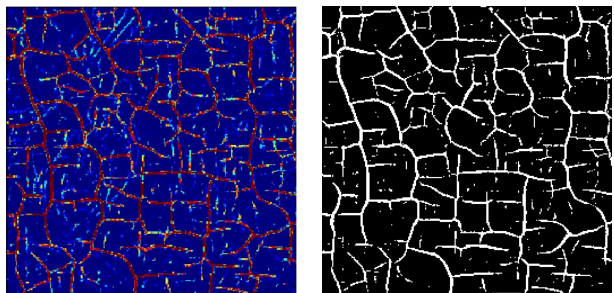

(a) Probability map (ranges from $P = 0$ (blue) to $P = 1$ (red))

(b) Crack map

Fig. 4: Output of the BCTF classifier

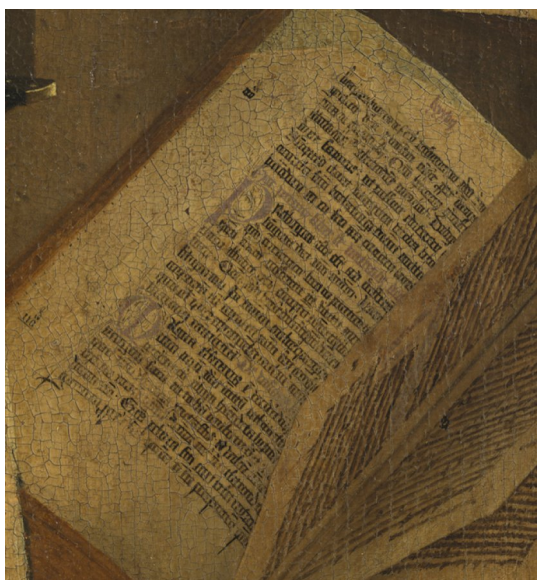

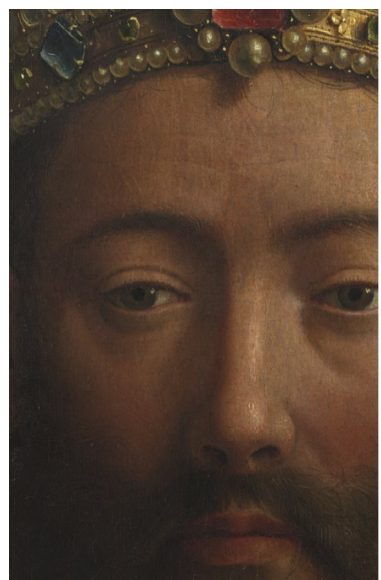

(a) De Visione Dei book

(b) Central figure

Fig. 5: Dataset

detection of all cracks. The main challenge consists of correctly detecting cracks without labeling the letters in the book as being cracks.

Another image consists of the face of the central figure in the Altarpiece whose identity has led to much discussion among scholars. Theories include that it is Christ, God the Father, or the Holy Trinity represented within a single person. The presence of other crack-like objects (e.g. letters in the book, eyelashes or hair, etc.) within these images makes both images an interesting selection for our experiments. Both images are depicted in Fig. 5.

## A. Training step

Accurately annotating crack pixels during the training phase can be quite challenging because of the nature of the cracks themselves. In this context the labeling is performed by using the semi-automated techniques described in [7] on arbitrarily chosen patches extracted from the area under investigation. The labeling of background pixels is done manually and is quite straightforward. Fig. 6 depicts the labeling for both crack- and non-crack pixels (in white) of the patch shown in Fig. 1. Note that not all pixels in the image are annotated, only the pixels of which we are certain. A possible issue that arises when working with more than one modality is the fact that some cracks might appear in one modality while being invisible in another. This happens when cracks or lacunas are covered (e.g. from a previous restoration) so that they are not visible at the infrared or visible wavelengths but appear in the X-rays. This may result in a somewhat flawed labeling of the training data and affects error rates when evaluating the classifier. This, and the lack of ground-truth data, are the primary reasons for which the classifier is assessed by means of visual results only.

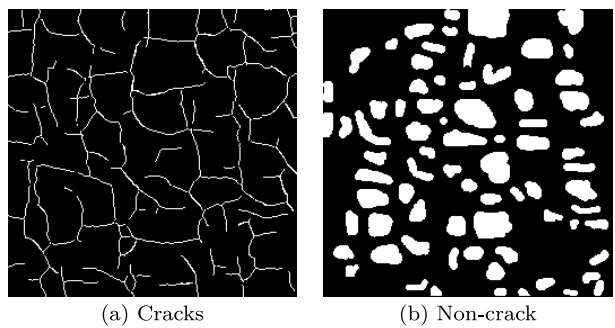

(a) Cracks  (b) Non-crack

Fig. 6: Selection of pixels for training (white means pixel is selected)

*B. Results*

*1) De Visione Dei book:* The image depicted in Fig. 5a has dimensions of approximately $3700 \times 3700$ pixels. The total number of annotated pixels used for training $n \approx 22100$ and the number of predictors $p = 208$ are all categorized to $d = 11$ bins. The important predictors $X_j$ that were selected by the BCTF method (together with their respective $k_j$ in brackets) are: $X_4(2)$, $X_{58}(2)$, $X_{132}(2)$, $X_{166}(2)$, $X_{196}(2)$, $X_{201}(3)$ and $X_{202}(3)$. Table I shows a more detailed description of the selected features. It can be observed that most of the features are taken from the X-ray domain. This makes sense since cracks in that part of the painting appear particularly clear in the X-ray. This is in strong contrast to the letters in the book, which are practically invisible (see Fig. 1) in the X-ray. The classification error for the BCTF is 0.03 while for RF it is 0.02. We ran the RF classifier with default parameters (i.e. 500 trees and 13 predictors were sampled for splitting at each node[3]). Fig. 7 shows the results for both our BCTF approach and the RF classifier on a selection from the book picture. From these figures it is clear that BCTF can easily compete with RF.

*2) Central figure:* The image is approximately $5400 \times 3800$ pixels in size. The number of annotated pixels is approximately 17200 and the number of predictors $p = 208$. Each predictor is categorized to $d = 11$ bins. BCTF again selected 7 important predictors: $X_5(2)$, $X_{34}(2)$, $X_{51}(2)$, $X_{58}(2)$, $X_{62}(3)$, $X_{79}(2)$ and $X_{196}(4)$. It is interesting to see that different features from different modalities were selected in comparison to the previous experiment (see Table II for a more detailed description of the selected features). The

---

[3]Software package can be found on:
http://code.google.com/p/randomforest-matlab/

TABLE I: Selected features for book image

| $X_j$ | $k_j$ | Description |
|---|---|---|
| $X_4$ | 2 | IR: Median filter (size filter: $3 \times 3$) |
| $X_{58}$ | 2 | IR: Elongated filter ($\sigma = 1$) |
| $X_{132}$ | 2 | Vis: Black Top Hat (size structuring element: $3 \times 3$) |
| $X_{166}$ | 2 | X-ray: Leung-Malik filter: directional |
| $X_{196}$ | 2 | X-ray: Elongated filter ($\sigma = 1$) |
| $X_{201}$ | 3 | X-ray: Frangi vesselness filter (vessel measure) |
| $X_{202}$ | 3 | X-ray: Frangi vesselness filter (vessel scale) |

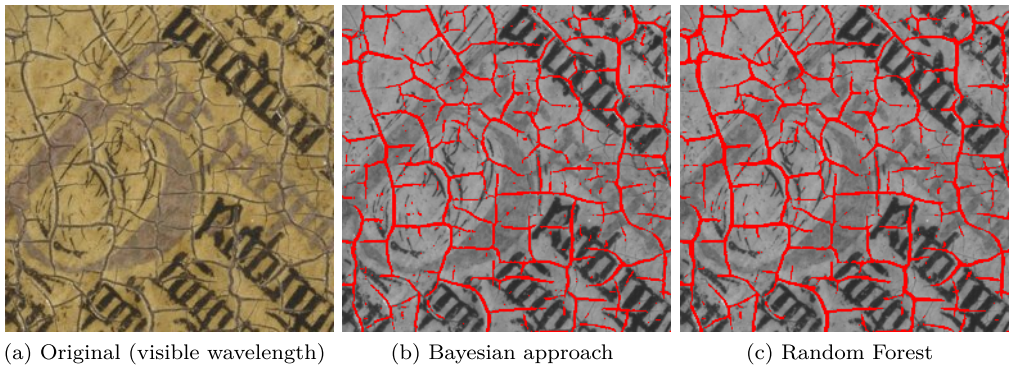

(a) Original (visible wavelength)     (b) Bayesian approach     (c) Random Forest

Fig. 7: Classification results: crack maps (marked in red) are overlaid on a grayscale version of the visible photograph

first five features are all taken from the infrared photographies, one is from the visible photography and the last corresponds to a feature in the X-ray domain. This is what we expected since the X-ray in that part of the painting is a lot less clear and contrasted than the X-ray of the book. It is also worth mentioning that the feature, corresponding to the result with elongated and directional filters, is favored in both experiments and for two different modalities (i.e. the infrared photography and X-ray). Fig. 8 shows classification results for both the BCTF approach and the RF classifier. Again, results are very similar although BCTF was able to detect more cracks in the eye when compared to RF.

TABLE II: Selected features for Central Figure image

| $X_j$ | $k_j$ | Description |
|---|---|---|
| $X_5$ | 2 | IR: Median filter (size filter: $6 \times 6$) |
| $X_{34}$ | 2 | IR: Leung-Malik filter: directional |
| $X_{51}$ | 2 | IR: Leung-Malik filter: LoG |
| $X_{58}$ | 2 | IR: Elongated filter ($\sigma = 1$) |
| $X_{62}$ | 3 | IR: Black Top Hat (size structuring element: $3 \times 3$) |
| $X_{79}$ | 2 | VIS: LoG ($\sigma = 5$) |
| $X_{196}$ | 4 | X-ray: Elongated filter ($\sigma = 1$) |

Fig. 9 shows the classification results when using BCTF for a crop where a lot of different objects are present. These results show the strength of our method, i.e. BCTF is still capable of correctly marking cracks, even in very busy areas, often with bad contrast between the cracks and other objects within the painting.

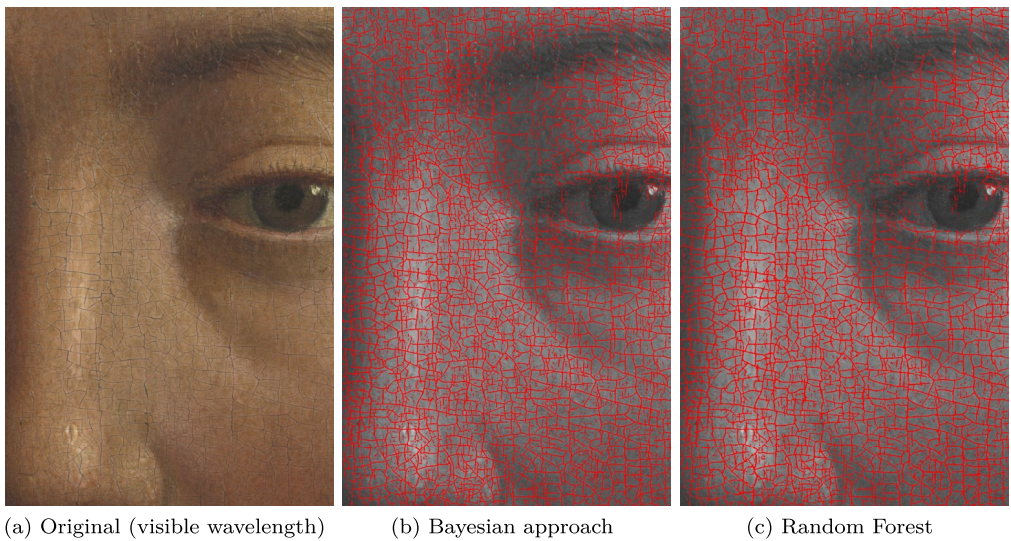

(a) Original (visible wavelength)     (b) Bayesian approach     (c) Random Forest

Fig. 8: Classification results: crack maps (marked in red) are overlaid on a grayscale version of the visible photograph

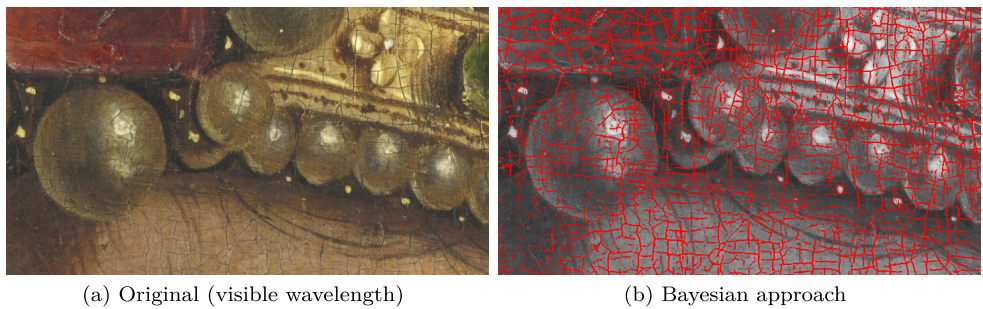

(a) Original (visible wavelength)     (b) Bayesian approach

Fig. 9: Classification results: crack maps (marked in red) are overlaid on a grayscale version of the visible photograph

## V. Conclusions and future work

In this paper we have introduced a novel crack detection method based on a nonparametric Bayesian conditional tensor factorization (BCTF) for high resolution images of paintings, coming from very diverse modalities. The proposed method has the appealing property that it is nonparametric and, once the most important predictors are selected, is faster than other state-of-the art methods such as the random forest classifier. Our method also shows that different modalities are beneficial for an accurate crack detection, as they provide more information than a single modality. BCTF is able to combine the information from the different modalities, unlike the methods proposed in [7], which were designed to work well on one single modality, namely the only photographic material of the Ghent Altarpiece available at that time. BCTF also confirms that the elongated directional filters, developed in [7] are a very strong feature to characterize cracks.

Future work includes making the data labeling process easier and more interactive. Also, since the labeled data is somewhat flawed, a more suitable error metric than classification error rate should be thought of in order to have a better objective assessment of the BCTF classifier. We further plan to extend the binary classification problem to a multi-class problem where new classes, such as different types of cracks and zones of restoration, will be introduced.

ACKNOWLEDGMENT

This research was supported by the Fund for Scientific Research Flanders (the Ph. D. fellowship and travel grant for long stay abroad V448712N of Bruno Cornelis).